\documentclass[letterpaper, 10 pt, journal, twoside]{ieeetran} 

\IEEEoverridecommandlockouts                              

\usepackage{color}
\usepackage{graphics} 
\usepackage{epsfig} 
\usepackage{mathptmx} 
\usepackage{amsmath} 
\usepackage{subfigure}
\usepackage[english]{babel}
\usepackage{graphicx}
\usepackage{makecell}
\usepackage{sidecap}
\usepackage{amsmath}
\usepackage{multirow}
\usepackage{amssymb}
\usepackage{hyperref}
\usepackage{algorithm}
\usepackage{algorithmicx}
\usepackage{algpseudocode}
\usepackage{enumitem}

\title{ForMIC: Foraging via Multiagent RL with Implicit Communication}

\author{Samuel Shaw$^{1,\dagger}$, Emerson Wenzel$^{1}$, Alexis Walker$^1$, Guillaume Sartoretti$^{2}$

\thanks{${\dagger}$ Corresponding author, to whom correspondence should be addressed.}
\thanks{Manuscript received: September, 9, 2021; Revised December, 20, 2021; Accepted February, 8, 2022.}
\thanks{This paper was recommended for publication by Editor M. Ani Hsieh
upon evaluation of the Associate Editor and Reviewers' comments.}
\thanks{$^{1}$ S. Shaw, E. Wenzel, and A. Walker are with the department of Computer Science at Tufts University, Medford, MA 02155, USA. {\tt\small \{samuel.shaw, emerson.wenzel, alexis.walker\}@tufts.edu}.}
\thanks{$^{2}$ G. Sartoretti is with the department of Mechanical Engineering at the National University of Singapore, 117575 Singapore. {\tt\small mpegas@nus.edu.sg}}%
} 

\begin{document}

\maketitle
\begin{abstract}

Multi-agent foraging (MAF) involves distributing a team of agents to search an environment and extract resources from it. Nature provides several examples of highly effective foragers, where individuals within the foraging collective use biological markers (e.g., pheromones) to communicate critical information to others via the environment. In this work, we propose ForMIC, a distributed reinforcement learning MAF approach that endows agents with implicit communication abilities via their shared environment. However, learning efficient policies with stigmergic interactions is highly nontrivial, since agents need to perform well to send each other useful signals, but also need to sense others' signals to perform well. In this work, we develop several key learning techniques for training policies with stigmergic interactions, where such a circular dependency is present. By relying on clever curriculum learning design, action filtering, and the introduction of non-learning agents to increase the agent density at training time at low computational cost, we develop a minimal learning framework that leads to the stable training of efficient stigmergic policies. We present simulation results which demonstrate that our learned policy outperforms existing state-of-the-art MAF algorithms in a set of experiments that vary team size, number and placement of resources, and key environmental dynamics not seen at training time.

\end{abstract}


\section{Introduction}
\label{RAL2022-introduction}

\noindent Multi-agent foraging (MAF) requires distributing a team of agents to explore an area and gather resources~\cite{zedadra2017multi}.
In the context of robotics, foraging systems could be applied to autonomous mining or harvesting.
Effective MAF approaches balance the \textit{exploration} for new resources with the \textit{exploitation} of already-discovered ones~\cite{yogeswaran2013reinforcement}, and enable the individual foragers to communicate key information to others (e.g., the discovery of a resource)~\cite{beckers1989colony,jackson2004trail}.
Many existing MAF algorithms take inspiration from social insects (e.g., bees, ants, and termites), which are highly effective foragers~\cite{beckers1989colony,jackson2004trail,greenleaf2007bee}, and use a combination of pheromones for communication and odometry for search and navigation.
In these works, while searching and navigating back to the nest, agents can leverage pheromone trails to make individual decisions by sensing the local pheromone concentration or gradient~\cite{zedadra2016cooperative,zedadra2014s,hecker2015exploiting,panait2004pheromone}.
However, such rule-based approaches are nontrivial to hand-craft and are usually suboptimal, since they depend heavily on the intuition and experience of the designers.
This is especially true in cases involving numerous resources, where pheromone trails can overlap and become more difficult to interpret.

\begin{figure}[t]
\vspace{0.12cm}
\begin{center}
\includegraphics[width=\linewidth]{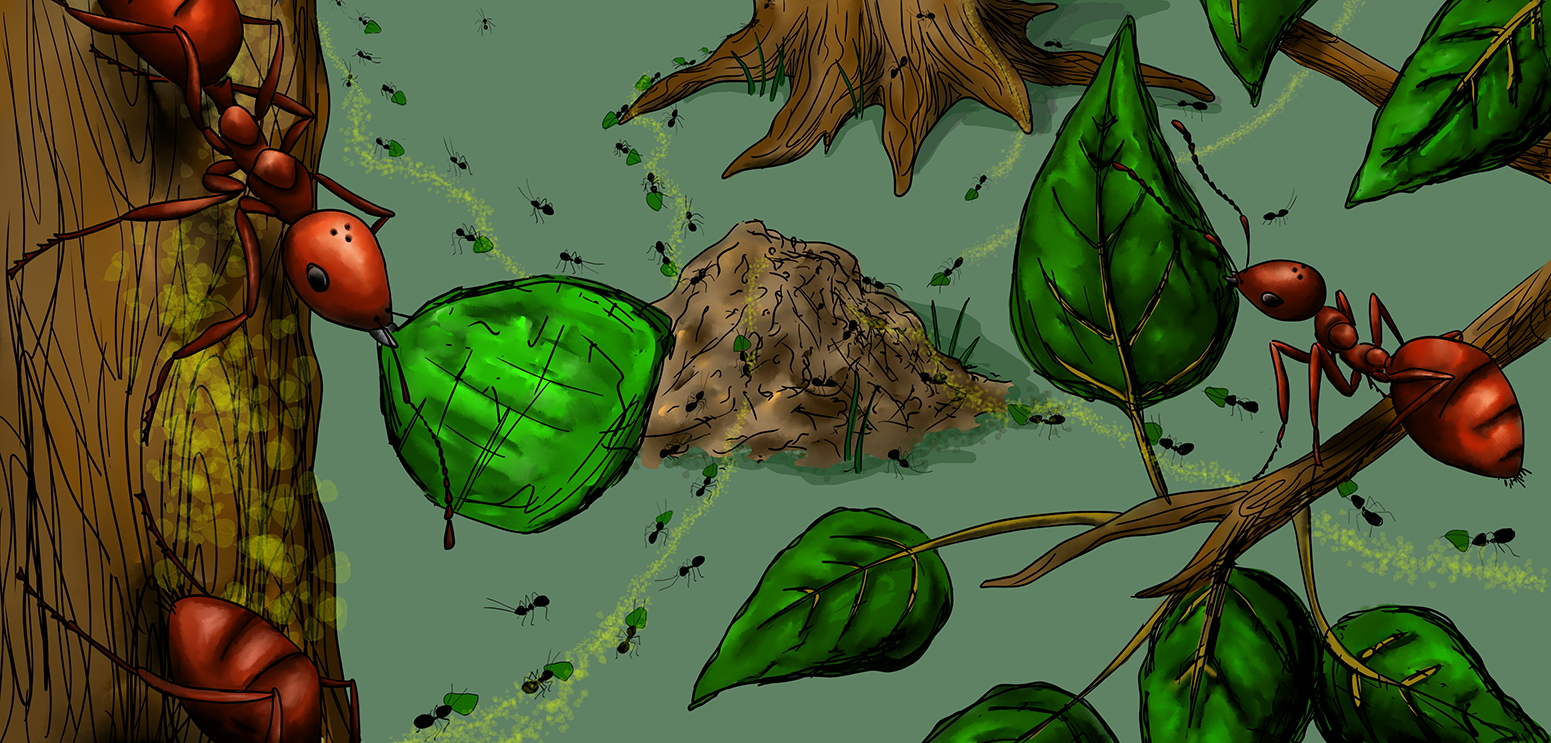}
\end{center}
\vspace{-0.55cm}
\caption{Many ant species use pheromones as a signaling mechanism to help with search and navigation. Colony members can make individual decisions based on the locally sensed pheromone concentration. Artwork by and courtesy of Anna Sz{\"u}cs.}
\label{RAL2022-fig:cover_letter}
\vspace{-0.55cm}
\end{figure}

In this paper, we propose a new avenue for pheromone-based MAF, \textit{ForMIC}, which extends previous works on distributed multi-agent reinforcement learning (MARL) of decentralized policies~\cite{sartoretti2019primal,tolstaya2020learning}. We endow agents with implicit communication abilities via \textit{stigmergy}: agent interactions via modifications of their common surroundings.
In our approach, agents can communicate effectively, yet in an \textit{implicit} manner, by learning how to interpret the locally sensed levels of pheromones placed by other agents.
We believe a learning-based MAF approach is beneficial for several reasons.
First, deep RL approaches, and in particular MARL algorithms, have been shown to naturally balance short- and long-term goals~\cite{yogeswaran2013reinforcement}, such as the exploration-exploitation trade-off in foraging tasks.
Second, learned approaches have been shown to be able to extract complex patterns from data~\cite{simonyan2014very,mnih2016asynchronous}, such as those present in the locally sensed pheromone concentrations.
Third, learned approaches are usually suitable for real-time use (fast forward inference), which is essential in the presence of highly dynamic environments that call for reactive planning.

However, training MAF policies with pheromones is highly nontrivial: agents need to learn how to interpret pheromones to navigate effectively, but also need to navigate effectively to place pheromones well.
In practice, this key issue often leads agents to simply ignore pheromone trails, since they are placed haphazardly by agents early on during training.
As such, we had to identify and develop a minimal set of key learning techniques that improve the stability of training and the performance and scalability of the final policy.
First, we devise a pheromone curriculum, which simplifies the learning task early on by providing agents with high-quality pheromone trails that they learn to trust.
Second, during training, we selectively restrict the action space of loaded agents to guide the learning of path planning for these agents, helping them to place better pheromone trails.
Finally, we introduce the concept of \textit{non-learning agents}, which enact the currently learned policy and allow us to drastically augment the team size during training at a lesser computational cost.

We train ForMIC's decentralized policy off-line in simulation and benchmark our approach by comparing our learned policy to existing, state-of-the-art, decentralized foraging approaches~\cite{zedadra2016cooperative,hoff2010two}, as well as a planner with full observability and centralized agent allocation.
ForMIC outperforms the benchmark strategies and is even comparable to this centralized planner in obstacle-free environments with infinite-capacity resources, while remaining amenable to real-time execution.
We further demonstrate that the reactive nature of ForMIC enables agents to forage in the presence of certain unmodeled (not seen during training) environmental dynamics.


\section{Prior Work}
\label{RAL2022-priorwork}


\subsection{Multi-Agent Foraging (MAF)}
\label{RAL2022-priorwork-MAF}

\noindent MAF is a widely studied problem in robotics and swarm intelligence~\cite{zedadra2017multi}.
The presence of numerous subtasks -- agent distribution/allocation, search, path planning, localization -- and the constant need to balance exploitation of known resources and exploration for new ones makes MAF particularly challenging.
Many existing approaches only focus on a subset of these subtasks, or propose fully observable and/or centralized solutions that do not scale well to larger teams~\cite{zedadra2017multi}.
For example, some works rely on a simplified version of the problem where there is only a single static resource available, moderating challenges related to agent distribution and negating the need for exploration once the resource is found.
Prior work has also studied single-resource MAF by allowing agents to serve as static ``beacons'' that help guide other agents between resources and nest~\cite{hoff2010two}.

Extending this idea of agents guiding other agents, and inspired by the collective intelligence of insect colonies, many recent works have also proposed models where robots automatically release virtual signals (i.e., pheromones) as they move, whose gradients can be measured locally to influence their decisions~\cite{song2020novel,zedadra2016cooperative,zedadra2015distributed,hecker2015exploiting,panait2004pheromone}.
Using pheromone trails to aid in navigation has been applied to single-resource MAF problems, dynamic single-resource MAF problems (where the one resource's location changes with time)~\cite{panait2004pheromone}, and more challenging multi-resource MAF problems.
Multi-resource MAF problems require increased emphasis on exploration and identification of resources and prior approaches have accordingly studied methods to tackle this.
Methods that use pheromones to signal search coverage~\cite{zedadra2016cooperative} and non-pheromone methods, such as geometric coverage or potential fields~\cite{magdy2013tornado}, have been studied to aid with exploration.
Other recent works have gone further and combined the use of pheromone trails and geometric coverage to improve upon the distributed exploration part of the MAF problem~\cite{zedadra2015distributed}.

Still other works have focused on the exploration/exploitation balance problem in MAF and have proposed general guidelines to switch from the exploitation of known resources to the exploration for more resources~\cite{hecker2015exploiting}.
In particular, notable recent works by Zedadra et al. have studied the automatic switching between individual and collective foraging behaviors~\cite{zedadra2016cooperative}, showing improved performances over a large spectrum of other pheromone-based approaches.

Significant research effort has also been devoted to replicating the dynamics of biological pheromones on physical robots, to deploy pheromone-based algorithms.
A variety of avenues for physical markers have been studied: alcohol/evaporation-based~\cite{pfeifer1998simulated}, heat~\cite{russell1997heat}, RFID~\cite{mamei2005physical}, and visual~\cite{svennebring2004building}.


\subsection{Communication in MARL}
\label{RAL2022-priorwork-MARLcomms}

Allowing agents to communicate can let them make better-informed local decisions or even cooperatively select joint actions, thus improving the team's performance.
To this end, the community has recently proposed approaches where agents learn how and what to communicate among the team, either by relying on reinforcement learning~\cite{jiang2018learning,foerster2016learning,mordatch2018emergence}, backpropagation through the communication channel~\cite{RAL2020-MARL_comms,das2019tarmac,sukhbaatar2016learning}, or via a form of shared memory~\cite{peng2017multiagent}.
In particular, some more recent works have leveraged the (learned) ``message passing'' algorithm at the core of Graph Neural Networks (GNNs) to endow agents with local communication abilities in systems with dynamic communication graphs/topologies~\cite{tolstaya2020learning}.
However, most of these methods still assume ``perfect'' (and therefore unrealistic) communication among agents (i.e., noise-free communication channels with no limit over bandwidth).
Further, only a few works to date have addressed the need for agents to select whom to communicate to/from, or when to even exchange information, to improve cooperation while minimizing the communication burden~\cite{jiang2018learning,das2019tarmac,priya2020neurosymbolic,paulos2019decentralization,kim2018learning}.
As a result, many existing works scale poorly to larger teams, where communications can either be overwhelming or insufficient to ensure agents receive and identify the right information for their decisions.


\section{Problem Formulation and Representation}
\label{RAL2022-problemFormulation}

\noindent In this section, we describe the specific MAF problem and the pheromone model considered in this work.


\subsection{Foraging Environment}
\label{RAL2022-problemFormulation-foragingEnvironment}

We consider a two-dimensional, grid-world environment, with randomly placed resources and obstacles (initially unknown to the agents) and a single, central nest.
Agents can collect food from resources and deposit it at the nest.
The nest has four cache locations, allowing four agents to deposit at the nest simultaneously. 
In order to keep these agent-resource/nest interactions realistic, agents must wait their turn in a queue before harvesting or depositing food.
When an agent interacts with a resource or the nest, the \textit{gathering rate} or \textit{dropoff rate} ($0 \leq rates \leq 1$) determines how long it takes for the agent to reach or empty its one-unit capacity.
These rates, and the number of nest entries and resources, can be adjusted to define various foraging tasks (e.g., a mining task might be more time-intensive than other collection-related tasks).


\subsection{Pheromone Dynamics}
\label{RAL2022-problemFormulation-pheromoneDynamics}

In our formulation of MAF, agents carrying food (i.e., full
agents) \hfill automatically \hfill leave \hfill a \hfill trail \hfill of \hfill pheromones \hfill on \hfill the
\newpage \noindent
ground, which other agents can (learn to) use to make contextual movement decisions.
These pheromone trails act as a form of short-term, collective memory that provides insight into the recent movements of loaded agents (e.g., which direction a loaded agent came from).
Although these pheromones decay with time, agents can reinforce this memory by repeating the same sequence of actions (i.e., collecting from the same resource again) or alter it by performing a new sequence. 


We represent pheromone concentration from agent $k$ at position $(i,j)$ at time $t$ as $p_t(i,j,k) \in [0,1]$, where a value of 1 indicates pheromones of maximum concentration.
The concentration of the pheromones left by a full agent decreases exponentially with the number of steps the agent has taken since harvesting from a resource:

\vspace{-.35cm}
\begin{equation}
    p_t(i,j,k) = \alpha^{t - h},
\end{equation}
\vspace{-.5cm}

\noindent where $\alpha \in [0,1]$ (we choose $\alpha = 0.97$) is the action decay rate, and $h \in \mathbb{W}$ is the time step at which the agent finished harvesting the resource.
This ensures that if an agent becomes lost, its pheromones will lose intensity, preventing it from misdirecting other agents. 

At a specific location, $(i,j)$, the pheromones from one agent can overpower those of another: the highest concentration pheromones dominate. 
Therefore, the total pheromone concentration at position $(i,j)$ at time $t$ is:

\vspace{-.3cm}
\begin{equation}
    P_t(i,j) = \max_k~ p_t(i,j,k).
\end{equation}
\vspace{-.4cm}

Additionally, unless refreshed by agent traffic, pheromones decay exponentially with time to avoid accumulation of pheromones and to allow agents to ``forget'' about resources that are too far from the nest or depleted~\cite{zedadra2014s}:

\vspace{-.3cm}
\begin{equation}
    P_t(i,j) = \beta \cdot P_{t-1}(i,j),
\end{equation}
\vspace{-.45cm}

\noindent where $\beta \in [0,1]$ is the decay rate (in practice, we choose $\beta = 0.99$).
Higher values of $\beta$ allow more persistent trails, which help agents to signal more distant food sources to each other.

The relative magnitudes of $\alpha$ and $\beta$ alter the pheromone dynamics.
Specifically, if $\alpha < \beta$, the pheromone trail will have a decreasing gradient with agent movement. 
This gradient reverses when $\alpha > \beta$ and is non-existent (i.e., the entire trail remains at the same level as it decays) when $\alpha = \beta$.

\begin{figure}
\vspace{-0.2cm}
\begin{center}
\includegraphics[width=\linewidth]{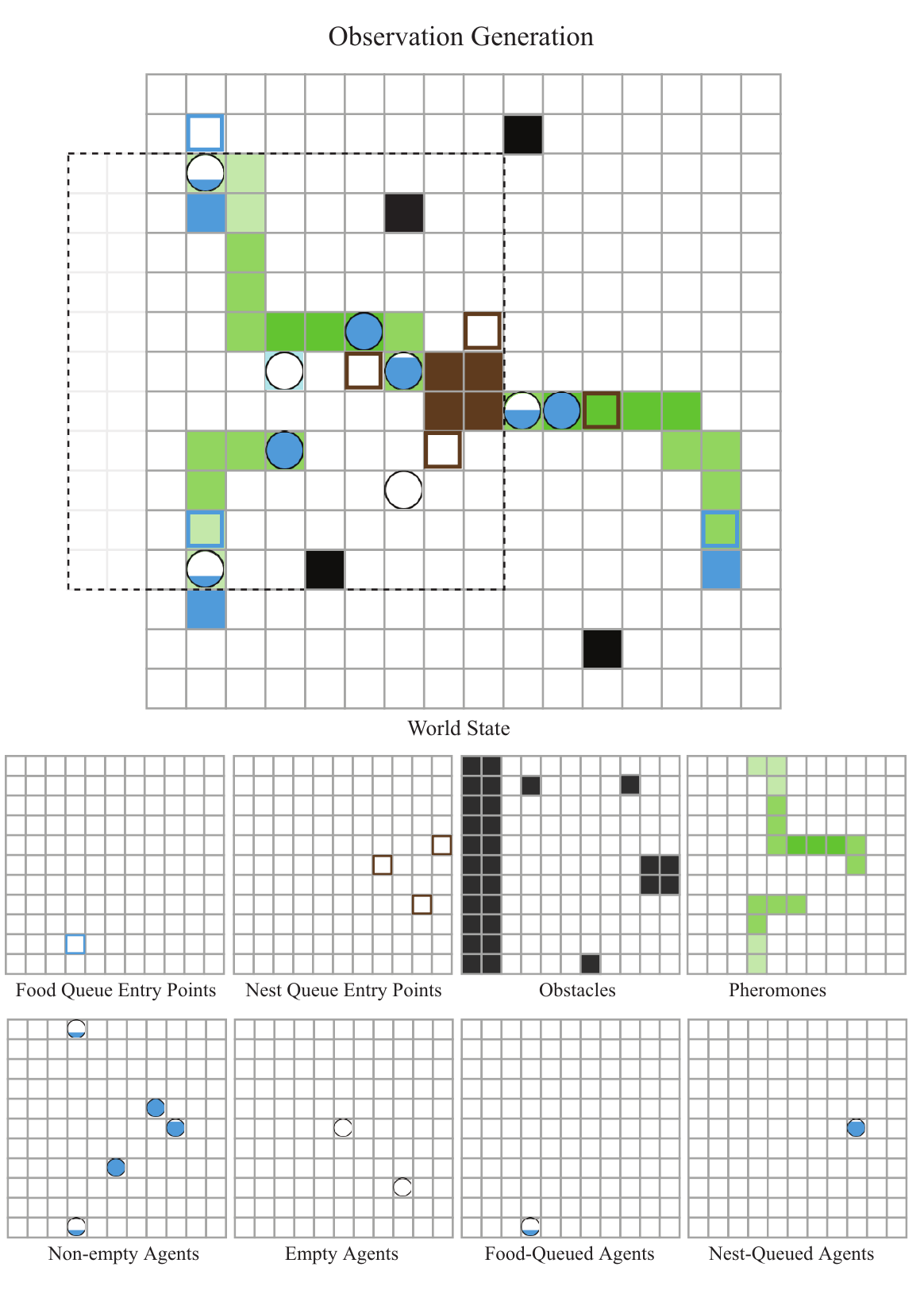}
\end{center}
\vspace{-.7cm}
\caption{Visual representation of a $16 \times 16$ world (top) and an agent's eight observation channels extracted from its $11 \times 11$ field-of-view, outlined with a black, dashed line (bottom). 
The resources and the nest are shown as fully colored blue and brown squares, respectively; queue entry points for each are shown as a square outlined in the corresponding color.
Obstacles are shown as fully colored black squares.
Agents are shown as circles filled proportionally to the amount of carried food (i.e., a circle colored entirely blue represents a full agent).
Pheromones are shown in green, such that lower-intensity pheromones appear more transparent.
}
\label{RAL2022-fig:obs}
\vspace{-0.4cm}
\end{figure}


\section{Policy Representation}
\label{RAL2022-policy}

\noindent In this section, we explain how we cast this foraging task in the RL framework by detailing the agents' observation and action spaces, and outlining the reward structure.   


\subsection{Observation Space}
\label{RAL2022-policy-observationSpace}

In this work, matrical data captures information about the statuses of cells surrounding the agent within a finite, square-shaped field-of-view (FOV) centered at the position of the agent.
For our implementation, we choose a fixed FOV of $11 \times 11$.
In general, the FOV magnitude dictates how far an agent should observe relative to its size.
To aid in the agents' learning, we partition the information within an agent's FOV into different channels, as is shown in Figure~\ref{RAL2022-fig:obs}.
Ultimately, an agent's observation is an $11 \times 11 \times 8$ tensor containing the matrical data that we provide as input to the network.

\begin{figure*}[t]
\begin{center}
\includegraphics[width=\linewidth]{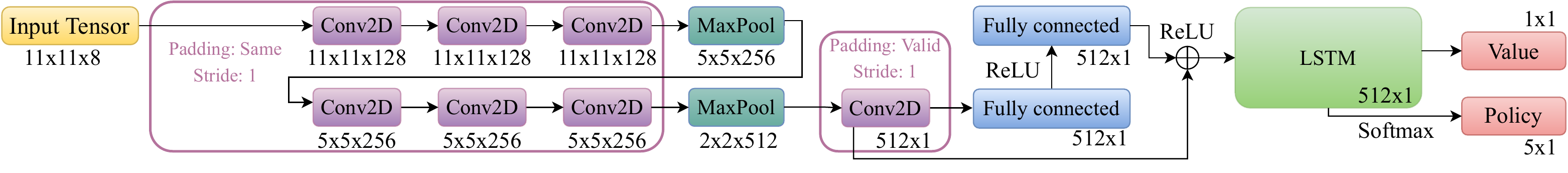}
\end{center}
\vspace{-0.7cm}
\caption{Neural network used in this work. The input is formed of eight $11 \times 11$ 2D channels, encoding spatial information about the agent's surroundings. This input is processed by a series of convolutional and maxpooling layers before being fed into an LSTM, to finally obtain the policy and value outputs.}
\label{RAL2022-fig:net}
\vspace{-0.4cm}
\end{figure*}


\subsection{Action Space}
\label{RAL2022-policy-actionSpace}

We grant agents five discrete actions to navigate within the environment and interact with resources or the nest to harvest or deposit food, respectively.
Four of these actions provide movement in each of the four cardinal directions in the two-dimensional grid world; each results in a one-cell movement that takes one time step to execute.

The final action enables agents to join queues associated with resources or caches in the environment, if an agent is on or adjacent to the designated entry cell for the queue.
Although it takes only one time step for an agent to join a queue, an agent may have to wait for other agents that are ahead of it in line before it can interact with the nest/resource itself.
While an agent waits in the queue and interacts, it is unable to select any of the described actions; an agent must commit to the line that it joins, and is only free to select actions again once its interaction is complete.


\subsection{Reward Structure}
\label{RAL2022-policy-rewardStructure}

To help agents learn efficient routes between resources and the nest (i.e., bring back food as quickly as possible), we provide agents with a small negative penalty at each time step during which they move or wait in a queue.
When an agent successfully collects food from a resource or deposits it at the nest, we provide the agent with a large positive reward.
Finally, agents are provided with a moderate negative reward when they attempt to join the wrong type of queue (e.g., a full agent attempting to join a resource queue).
The reward values are shown in Table~\ref{RAL2022-reward-table}.

\begin{table}[H]
\vspace{-0.2cm}
\begin{center}
\begin{tabular}{|p{5cm}|p{1.2cm}|}
\hline
\textbf{Action} & \textbf{Reward}                                                                                      \\ \hline
Movement Step & \,-0.05                    \\ \hline
Queued Step  & \,-0.05                                   \\ \hline
Successful collect or deposit              & 10.00                  \\ \hline
Entered correct queue             & \,\,\,0.00                                \\ \hline
Attempt to enter incorrect queue              & \,-1.00 \\ \hline
\end{tabular}
\vspace{0.1cm}
\caption{ \label{RAL2022-reward-table} \vspace{-0.3cm}}
\end{center}
\vspace{-0.4cm}
\end{table}


\subsection{Actor-Critic Network}
\label{RAL2022-policy-ACNet}

This work takes inspiration from several sources for the choice of learning algorithm and the design of the policy network.
We rely on the extension of the asynchronous advantage actor-critic (A3C) learning algorithm~\cite{mnih2016asynchronous} that we developed in our previous work~\cite{sartoretti2019primal}.
In this algorithm, agents share a common learning environment and each agent has its own local copy (used for policy estimation, as well as other outputs described below) of a common (global) policy neural network.
At the end of each learning episode, each agent calculates local learning gradients based on its own experience and pushes these gradients to the global network.
These gradients are summed together and applied to the global network before each agent updates its local network with the new global weights.
Then, a new episode starts.

Our network structure is inspired by VGGnet~\cite{simonyan2014very}, and is shown in Figure~\ref{RAL2022-fig:net}.
The $3 \times 3$ convolutions of VGG are well-scaled for our relatively small, channeled input, which resembles the structure of a small image.
The LSTM cell is a critical addition, since it provides agents with implicit memory of their recent interactions and enables them to learn policies with temporal dependencies (e.g., odometry).
In our experimentation, we tried two other well-studied networks for image recognition: AlexNet and ResNet.
With both of these larger networks, we observed no performance improvement but a significant decrease in training speed.

During training, the policy and value outputs are updated in batch every $n = 512$ steps (i.e., at the end of each episode).
As is common in RL, the value output of the network is updated toward the total discounted return ($R_t=\sum_{i=0}^k\gamma^i r_{t+i}$) by minimizing:

\vspace{-.6cm}
\begin{equation}
L_{V}=\sum_{t=0}^T (V(o_t;\theta)-R_t)^2,
\label{RAL2022-valueloss}
\end{equation}
\vspace{-.2cm}

\noindent where $o_t$ is the input of an agent at time step $t$, and $\theta$ is the set of parameters (weights) of the neural network.

To update the policy distribution, we rely on an approximation of the ``advantage'' of the action selected in each state, $a_t$, using the state value function:
$A(o_t, a_t; \theta) = \sum_{i=0}^{k-1} \gamma^i r_{t+i} + \gamma^k V(o_{k+t}; \theta) - V(o_t; \theta)$ (where $k$ is bounded by the batch size $T$, and $\gamma$ is the discount factor). 
We also add an extra entropy term $H(\pi(o))$ to the policy loss, to encourage exploration and discourage premature convergence of the learning process by penalizing a policy that is too certain during early training:

\vspace{-.5cm}
\begin{equation}
\label{RAL2022-policyloss}
L_{\pi} = \sigma_H \cdot H(\pi(o)) - \sum_{t=0}^T \log(P(a_t | \pi, o; \theta) A(o_t, a_t; \theta)),
\end{equation}
\vspace{-.25cm}

\noindent with a small entropy weight $\sigma_H$ ($\sigma_H = 0.01$ in practice).
As is now common, we rely on a loss function, $L_{valid}$, to train agents to learn valid actions.
This loss minimizes the log likelihood of selecting invalid moves, thus decreasing their activation probability in the policy output.
The final training loss reads: $L = L_{\pi} + 0.5 \cdot L_{V} + 0.5 \cdot L_{valid}$.


\section{Learning}
\label{RAL2022-learning}

\noindent In this section, we describe the minimal set of learning techniques we developed and their function during training.
These techniques are crucial for dealing with the circular dependency associated with learning stigmergic interactions: agents need to learn how to interpret signals from other agents (here, pheromone concentrations around them) to navigate effectively, but also need to navigate effectively to place pheromones well themselves.


\subsection{Non-learning Agents}
\label{RAL2022-learning-nonlearning}

The team size (i.e., number of agents) is an important parameter to consider in MAF.
In small teams, a handful of resources -- or perhaps just one -- close to the nest may be enough to ensure every agent is meaningfully occupied.
Differently, in larger teams, resources close to the nest quickly become overwhelmed, forcing agents to seek out resources further away; these are the more interesting MAF scenarios. 
More generally, depending on the amount of ``traffic" that a resource sees and the resource harvesting rate, it may be advantageous for an agent to navigate to another resource to maximize its contribution to the team.

In this work, we propose to rely on \textit{non-learning agents} to allow us to train the policy in larger teams without increasing the computational cost.
Non-learning agents execute the currently learned policy based on their observations, since performing forward inference of their policy network is low cost.
Furthermore, as demonstrated in our comparison results (Section~\ref{RAL2022-modelComparisons}), the introduction of these non-learning agents also allows us to better leverage the distributed learning framework to achieve a scalable foraging approach.

We use team sizes of $8 + m$, where there are eight learning agents and $m$ non-learning agents.
We further propose to vary this number of non-learning agents across the four A3C meta-agents (independent environments in which independent teams train).
We use team sizes of $16$, $24$, $32$, and $48$ agents (n.b., each team is composed of a constant $8$ learning agents, while others are non-learning).


\subsection{Action Filtering}
\label{RAL2022-learning-actionExecution}

During training, we rely on the pruning of invalid/poor actions from the agents' action space; at testing time, we allow agents to freely select their actions.
More specifically, if an agent selects an invalid action, a new action is drawn instead at random from the collection of valid actions and enacted.
The additional loss function $L_{valid}$ also helps agents quickly learn which actions are valid.
We observed that the agents' valid action rate reaches $85\%$ within a few thousand episodes, before exceeding $95\%$ in the final policy ($\sim 24$k~episodes).

Further, to speed up training and increase the performance of the final policy, we designate actions as invalid for two reasons. 
As is commonly done, we consider an action to be invalid if it violates the environment dynamics: moving into an occupied cell, attempting to join a queue that does not exist, etc. 
In this work, we also use action validity to enforce hard constraints in the policy that can speed up and stabilize training.
More specifically, when an agent is full, actions that move the agent away from the nest (in Euclidean distance, using simple dot product) are considered invalid unless other actions are physically impossible.
Experimentally, we found that employing this constraint was essential for agents to both learn how to return to the nest and drop/utilize pheromones properly (see our ablation study in Section~\ref{RAL2022-modelComparisons}).

Additionally, we explored action filtering based on A* path length, which would better handle environments containing non-convex obstacles or dense clusters of agents, but found training to be unstable and slow. 
We believe that this instability could be because -- unlike in Euclidean distance action filtering -- small changes in the world outside an agent's observation window can drastically affect which actions are favorable. 
Computationally, filtering via A* is much more expensive, since it replaces simple dot products with A* calls for each agent at each time step.


\subsection{Pheromone Curriculum}
\label{RAL2022-learning-curriculum-pheromones}

Since learning to effectively use agent-placed pheromones is a challenging task, we also employ a pheromone curriculum that eases the learning process. 
A pheromone curriculum may also be looked at as a form of \textit{imitation learning}, where an agent learns from an expert demonstrator~\cite{montgomery2016guided}.
In early episodes when agents do not navigate well, we initialize the world with pheromone trails (which we term ``pheromone highways'') from every resource to the nest in a direct path.
These pheromone highways are initialized with the same pheromone gradient that agents would create (i.e., appear to be placed by agents) during the episode based on $\alpha$ and $\beta$ (see Section~\ref{RAL2022-problemFormulation-pheromoneDynamics}), but do not decay further during the episode.

We propose a pheromone curriculum to control the relative intensity of these highways, versus the (decaying) pheromone trails left by agents carrying food.
Specifically, the pheromone highways are initially at full strength (i.e., maximum pheromone concentration, a value of $1$) and agent pheromones have zero strength.
Between $5$k and $10$k training episodes, we linearly decrease the strength of the pheromone highways and simultaneously linearly increase that of agent pheromones.
At the $10$k episode mark, the pheromone highways are at zero strength (i.e., no longer visible to agents) and the agent pheromones are at full strength.
By controlling these intensities over time, we help agents steadily build trust in the pheromone trails and learn to deposit meaningful pheromones for each other.

In our comparison results (Section~\ref{RAL2022-modelComparisons}), we show that the pheromone curriculum is essential for learning.
Without it, the problem becomes too difficult too quickly and agents fail to learn an effective policy, or learn to disregard the pheromone trails and converge to a poorer policy.


\subsection{Learning Hyper-parameters}
\label{RAL2022-learning-parameters}

\begin{figure*}[t]
\begin{center}
\includegraphics[width=0.49\linewidth, height=0.23\linewidth, keepaspectratio=false]{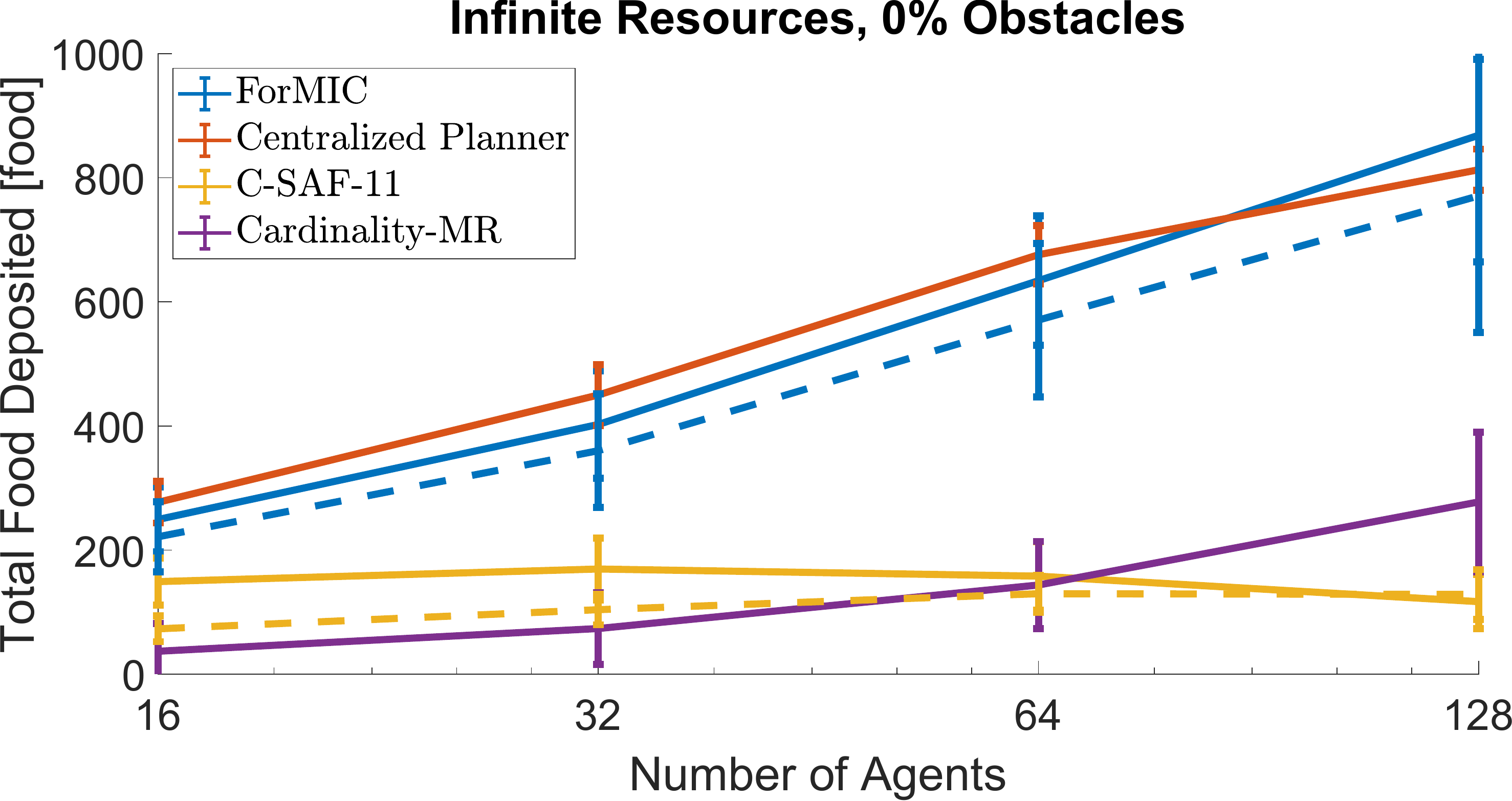} \hfill
\includegraphics[width=0.49\linewidth, height=0.23\linewidth, keepaspectratio=false]{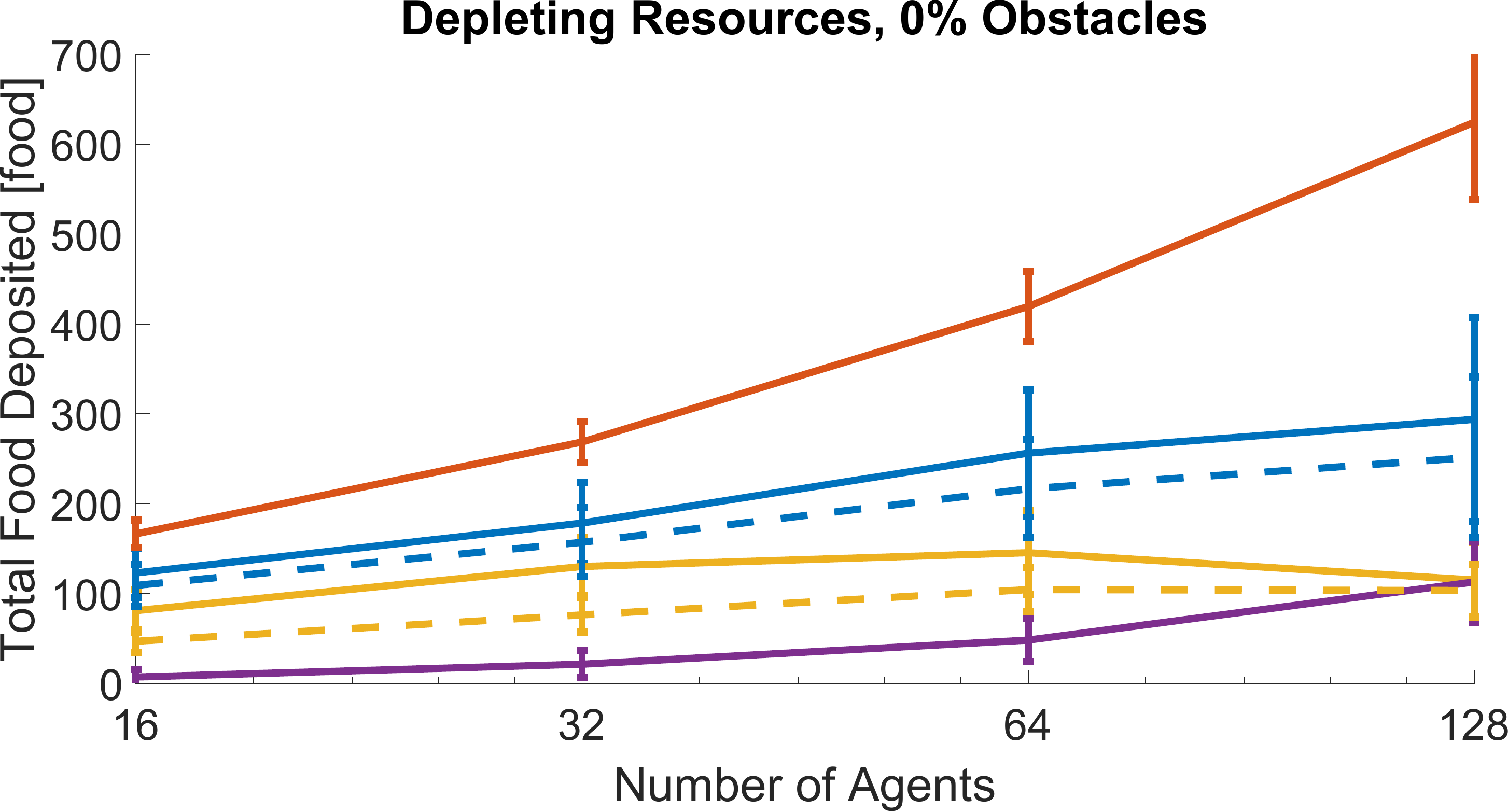}\\[0.15cm]
\includegraphics[width=0.49\linewidth, height=0.23\linewidth, keepaspectratio=false]{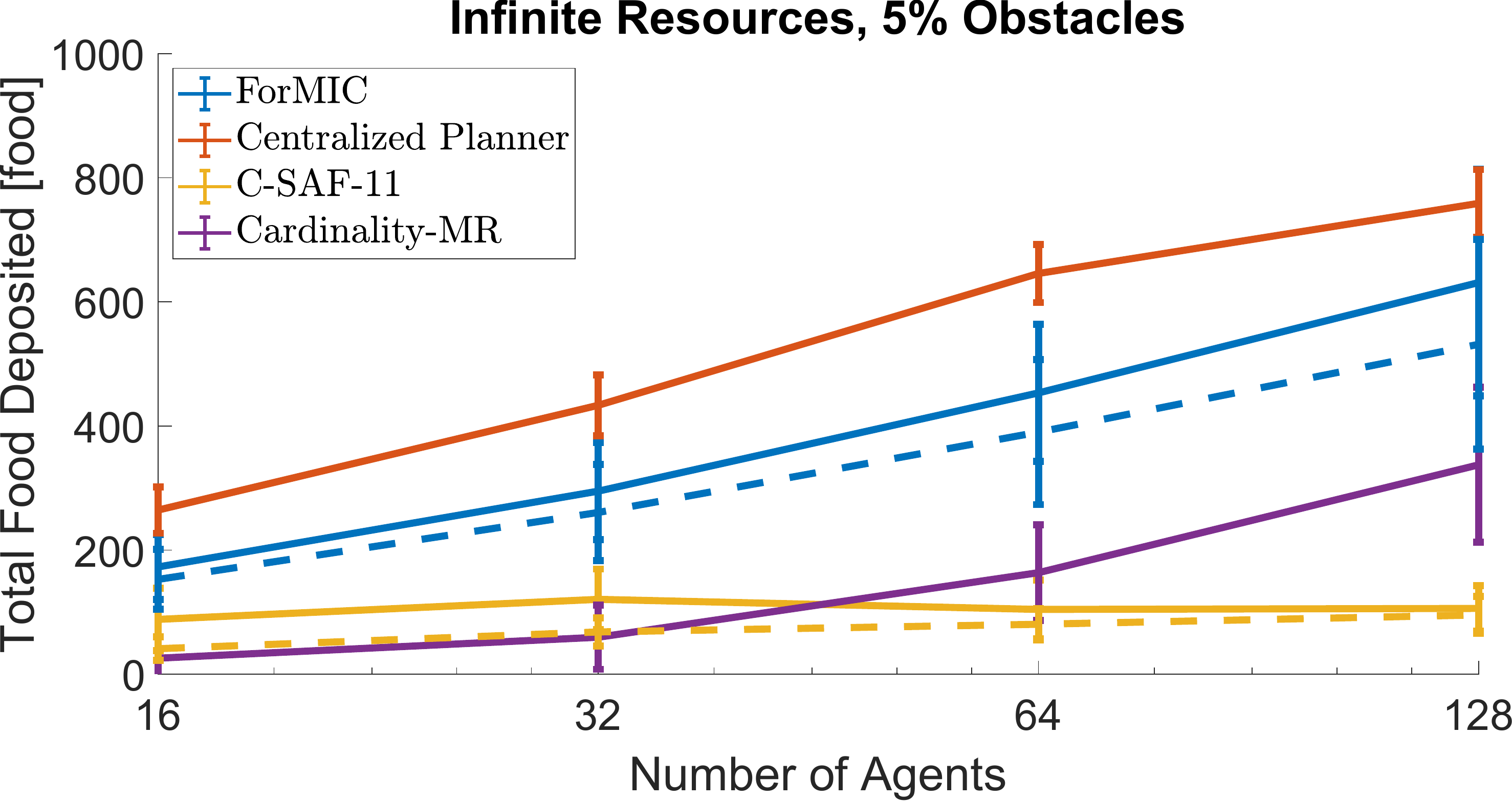} \hfill
\includegraphics[width=0.49\linewidth, height=0.23\linewidth, keepaspectratio=false]{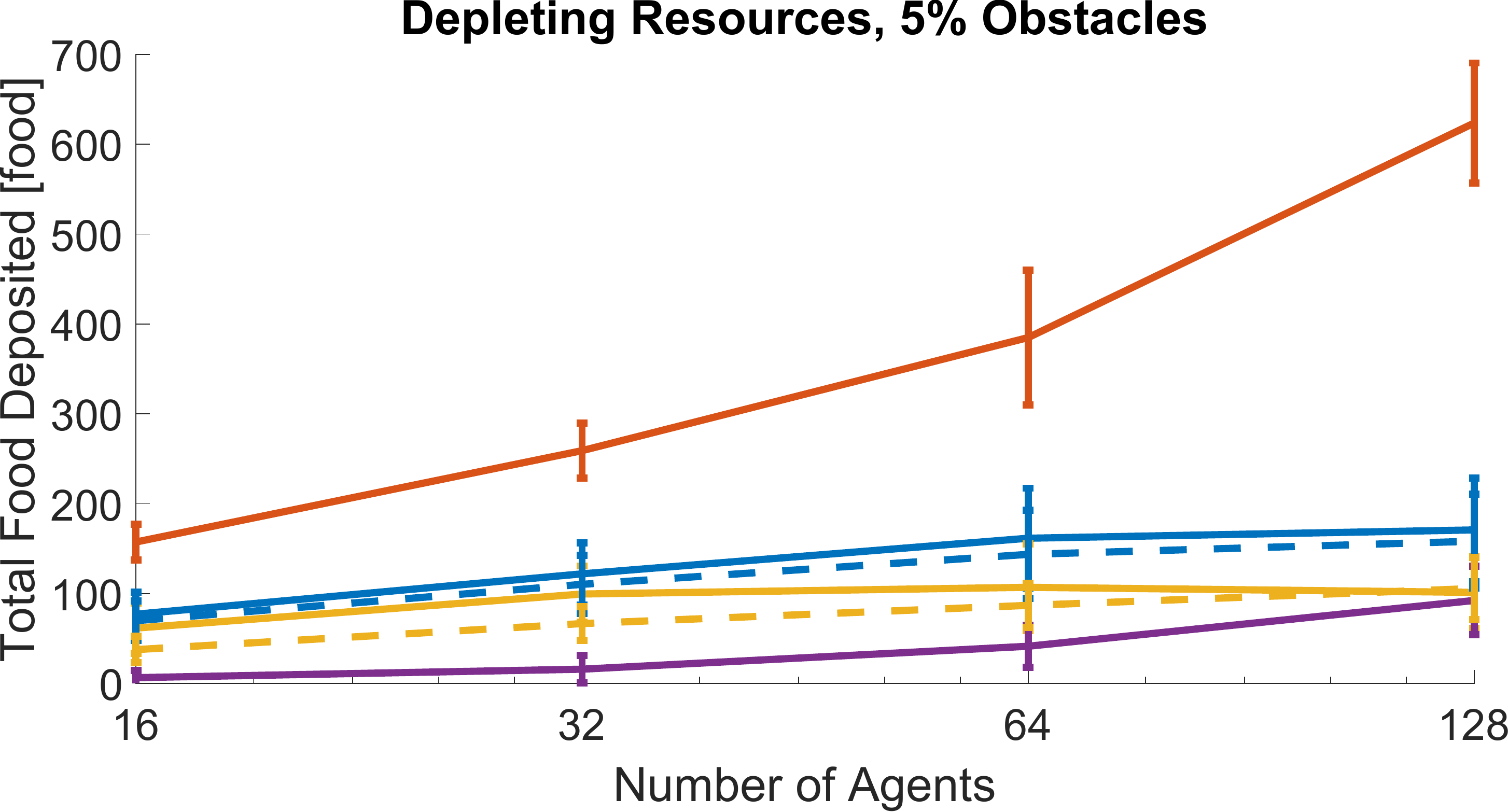}
\end{center}
\vspace{-0.55cm}
\caption{Comparison of algorithm throughputs (total food deposition at the nest) in experiments across a variety of team sizes. Approaches that use pheromones have an additional dashed curve that shows their performance in the presence of $3$ pheromone wipeouts of $100$ time steps each (i.e., instantaneous, sustained pheromone degradation).
ForMIC scales well and outperforms C-SAF-11 and Cardinality-MR for all large team sizes, with or without pheromone wipeouts.
In the best case (top left), ForMIC's performance mirrors that of the centralized planner with full observability of the world and global communications.
In other scenarios, with depleting resources or obstacles, the centralized planner is at an advantage: full observability enables it to plan more intelligent paths in obstacle-cluttered environments, and global communication enables it to never over-schedule resources.
We believe ForMIC's results are notable for two additional reasons: 1) the ForMIC policy was \textit{not} trained in scenarios featuring pheromone wipeouts or depleting resources, and 2) it can be run in real-time, unlike the centralized planner which is computationally burdensome and does not scale well with additional agents.
}
\label{RAL2022-fig:results}
\vspace{-0.4cm}
\end{figure*}

We train a single ForMIC model that we use for all test scenarios.
Training episodes last $512$ time steps, at the end of which we perform one gradient update for each agent.
We use the Nadam optimizer with a learning rate of $5 \cdot 10^{-6}$, and a discount factor $\gamma = 0.95$. 
We train in four independent environments (A3C meta-agents), synchronizing agents in the same environment at the beginning of each step and allowing them to act in parallel (in a random order at each time step).

Training was performed on a desktop computer equipped with an Intel i9-10980XE 18-core CPU, 64Gb of RAM, and two NVIDIA RTX 2080Ti (only one used for training), and lasted close to five days.
The full code used to train agents, as well as all result plots on varying world sizes (and the trained ForMIC model), can be found at~\url{https://bit.ly/ForMICcode}.


\section{Simulation Experiments}
\label{RAL2022-experiments}

\noindent In this section, we describe the algorithms we benchmark against and which environmental parameters we vary.


\subsection{Benchmark Algorithms}
\label{RAL2022-experiments-benchmarks}

To evaluate the performance of our learned foraging policy, we benchmark ForMIC against state-of-the-art foraging algorithms that employ hard-coded, rule-based strategies derived from domain knowledge: C-SAF~\cite{zedadra2016cooperative} and Cardinality~\cite{hoff2010two}.
Additionally, we test our learned foraging policy against a planner with full observability and global communication to provide an upper bound on performance.


\subsubsection{C-SAF}
\label{RAL2022-experiments-benchmarks-handcrafted}

C-SAF uses a pheromone model to communicate search coverage by the presence of pheromones and utilizes the pheromone concentration gradient created to navigate back to the nest.
For a fair comparison in our tests, we created an augmented version of C-SAF with an $11 \times 11$ FOV (referred to as ``C-SAF-11"); the original algorithm only allows agents to observe adjacent cells, equivalent to a $3 \times 3$ FOV.


\subsubsection{Cardinality}
\label{RAL2022-experiments-benchmarks-cardinality}

Cardinality employs a pheromone-free strategy, where agents themselves can act as navigation markers (referred to as beacons) for other agents. 
An agent serves as a beacon when there are fewer than two beacons in the agent's field-of-view, and simply stops moving from there onward to instead relay messages within the team to allow non-beacon agents to navigate between the nest and resources.
Although the original description of cardinality is designed for use in an environment with a single resource, we adapted the algorithm for the multi-resource case (referred to as ``Cardinality-MR"). 


\subsubsection{Centralized Planner with Full Observability}
\label{RAL2022-experiments-benchmarks-optimal}

We also compare against a planner with full observability and global communication (i.e., knowledge of all resource locations/capacities, and current destination of all agents).
This planner uses a centralized, greedy allocation strategy to assign agents to targets, and decentralized path planning, since computing a fully joint plan would be intractable.
Nevertheless, we believe that this planner provides a reasonable (near-optimal) upper bound for our MAF problem, since agents can observe the entire environment (no search for resources needed) and have access to the current intentions of all agents (i.e., which agents are en route to which resources).


\subsection{Simulation Results}
\label{RAL2022-experiments-scenarios}

In order to provide a fair comparison across each of the described algorithms, we generate a standardized set of test MAF instances.
Testing episodes last 1024 time steps in a $128 \times 128$ world (i.e., larger than any training worlds), with a variable number of agents: 16, 32, 64, and 128.
We also vary obstacle density: no obstacles and $5\%$ obstacles (upper bound of the training range).
For each possible set of parameters, we generate 50 scenarios with a varying number of resources randomly distributed within the environment, allowing us to evaluate the variability due to randomized resource placement.
We evaluate ForMIC in three types of environments, two of which were \textit{not} seen at training time:
\begin{enumerate}[leftmargin=0.47cm]
\item Environments with infinite-capacity resources, and normal pheromone dynamics (same as training).
\item Environments with finite-capacity resources and normal pheromone dynamics. There, when agents deplete a resource, a new resource appears, placed randomly within the world (i.e., constant number of resources in the world). This behavior forces agents to constantly find new resources to exploit, as existing ones get depleted. 
\item Environments with infinite-capacity resources, but where the pheromone trails are \textit{wiped out} at randomly drawn times during each episode. We consider three 100-step wipeouts that each start with an instantaneous removal of all pheromones in the world, followed by $100$ steps during which pheromones cannot be placed or seen. A wipeout significantly interferes with implicit communication and disturbs the collective's memory.
\end{enumerate}

\begin{figure}[t]
\vspace{0.2cm}
\begin{center}
\includegraphics[width=\linewidth]{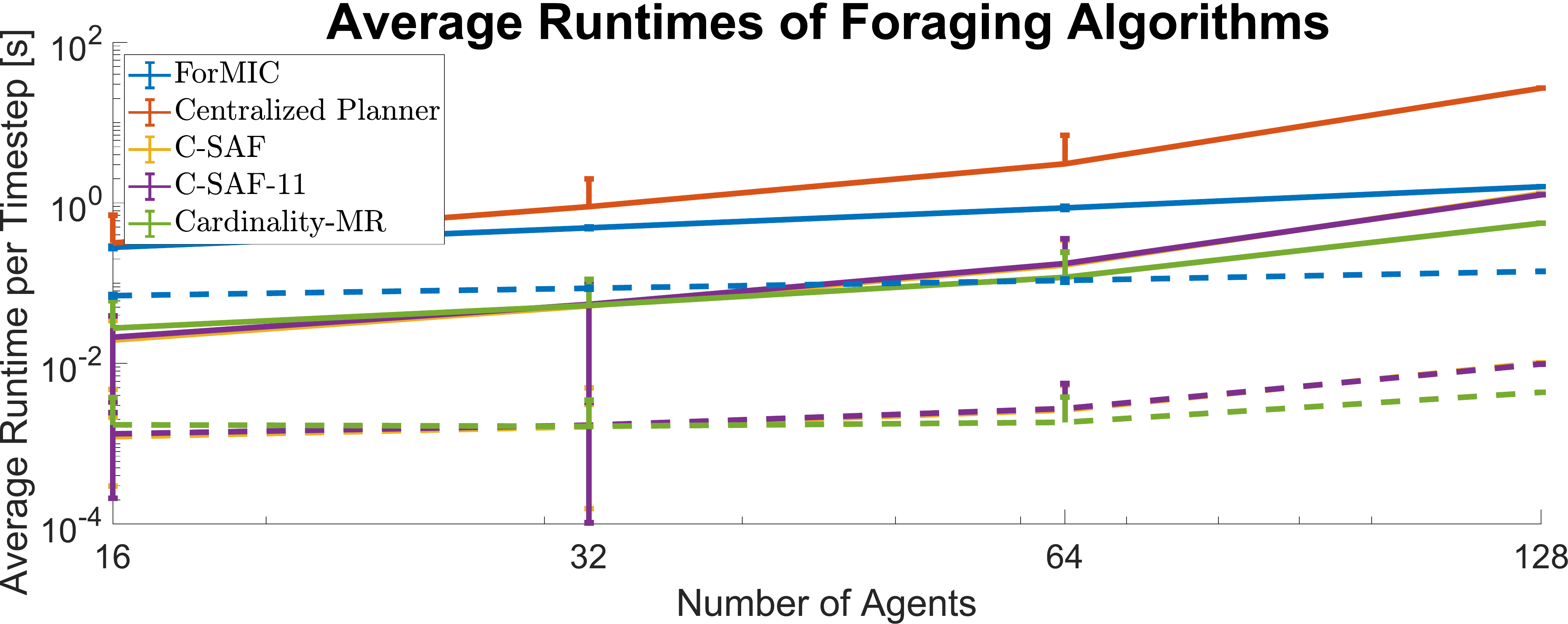}
\end{center}
\vspace{-0.5cm}
\caption{Comparison algorithm run times. Solid: measured run time per time step. Dotted: measured run time per time step redivided by team size to account for decentralized execution. Note the log scale on the vertical axis.}
\label{RAL2022-fig:run times}
\vspace{-0.4cm}
\end{figure}

\noindent We present summary results in Figure~\ref{RAL2022-fig:results}.
Figure~\ref{RAL2022-fig:run times} provides a comparison of the algorithms' run times.
Video results are available at \url{https://bit.ly/ForMICvideos}.


\subsection{ForMIC Models Comparisons}
\label{RAL2022-modelComparisons}

In the spirit of presenting the minimal, yet most effective MAF approach possible, we conducted an ablation study by training and comparing the performance of a variety of additional ForMIC variants, each with one of our core learning components removed.
Table~\ref{RAL2022-ablation-table} shows the relative performance of ForMIC variants compared to the highest-performing, full ForMIC model, averaged over all sets of experiments (i.e., with/without obstacles, with/without wipeouts, all team sizes), as described in Section~\ref{RAL2022-experiments-scenarios}.

\begin{table}[h]
\begin{center}
\vspace{0.4cm}
\begin{tabular}{|p{5.25cm}|p{2.25cm}|}
\hline
\textbf{Comparison Model}                  & \textbf{Relative Performance} \\ \hline
Training \& Testing without pheromones     & $82.79\% \pm 10.00\%$ \\ \hline
Training without pheromone curriculum      & $75.89\% \pm 7.00\%$ \\ \hline
Training without non-learning agents       & $72.58\% \pm 12.85\%$ \\ \hline
Testing without pheromones (empty channel) & $65.89\% \pm 8.73\%$ \\ \hline
Training without invalid action pruning    & $6.29\% \pm 1.98\%$ \\ \hline
\end{tabular}
\vspace{0.1cm}
\caption{Summary of our ablation study.}
\label{RAL2022-ablation-table}
\end{center}
\vspace{-1.2cm}
\end{table}
From this, we conclude that endowing agents with the ability to communicate implicitly via the pheromone channel improves the performance of the collective. 
Most importantly, we show significant performance improvement relative to a model trained with pheromones but tested without them, and a model trained and tested without pheromones.
Additionally, we show that our pheromone curriculum, non-learning agents, and invalid action pruning are all learning techniques that improve the performance of the trained model.

To further investigate the effects of non-learning agents, we trained several teams of 32 agents, varying the number of learning and non-learning agents. 
The results, summarized in Table~\ref{RAL2022-nonlearning-table}, demonstrate that similarly performant polices can be trained with less VRAM by substituting some learning agents for non-learning agents.
Although we had access to a NVIDIA Titan RTX (24GB VRAM) for final training, during development, we only had access to an older NVIDIA TITAN V (12GB) or 2080TI (11GB); introducing non-learning agents enabled us to develop ForMIC when we otherwise would not have been able to.

\begin{table}[h]
\begin{center}
\vspace{-0.1cm}
\begin{tabular}{|p{1.6cm}|p{1.6cm}|p{2.2cm}|p{1.6cm}|}
\hline
\textbf{Learning Agents} & \textbf{Non-learning Agents} & \textbf{Relative Performance} & \textbf{VRAM Needed}   \\ \hline
32 & 0 &$105.33\% \pm 10.88\%$ & $> 21$GB \\ \hline
16 & 16 &$94.16\% \pm 13.70\%$ & $> 12$GB \\ \hline
8 & 24 &$102.11\% \pm 19.23\%$ & $\leq 11$GB \\ \hline
\end{tabular}
\vspace{0.1cm}
\caption{Study of non-learning agents.}
\label{RAL2022-nonlearning-table}
\end{center}
\vspace{-0.8cm}
\end{table}
We also investigated the effects of varying the agents' FOV size, which requires retraining with slight changes to the network structure.
As would be expected and is summarized in Table~\ref{RAL2022-fov-table}, performance sharply declines as the FOV is restricted, but only improves marginally with a larger FOV.
\begin{table}[h]
\begin{center}
\vspace{-0.3cm}
\begin{tabular}{|p{2.00cm}|p{3.00cm}|}
\hline
\textbf{FOV Size}                  & \textbf{Relative Performance} \\ \hline
5  & $59.61\% \pm 11.00\%$ \\ \hline
21 & $105.79\% \pm 19.46\%$ \\ \hline
\end{tabular}
\vspace{0.1cm}
\caption{Study of Agent FOV Size.}
\label{RAL2022-fov-table}
\end{center}
\vspace{-0.8cm}
\end{table}

Finally, we tested our vanilla ForMIC model (no retraining) in scenarios involving different pheromone dynamics (i.e., out-of-distribution tests): 1) decaying trails with no gradient, 2) decaying trails with locally noisy gradient, 3) decaying trails with gradient-destroying noise (specifically, agents place pheromones with concentration uniform-randomly drawn between $0.2$ and $1$ at each time step).
Without any additional retraining, ForMIC performed identically on all these cases, showing no negative effect on its performance.
We believe that this indicates that the presence or absence of pheromones matters much more than the detailed pheromone content of each cell (which is drastically affected by noise in our experiments).
This may suggest that ForMIC could be easier to implement on robots in the real world, since its policy doesn't depend on a clearly formed pheromone gradient (or the existence of a gradient at all), providing a natural resilience against noise in the pheromone levels.

\section{Discussion}
\label{RAL2022-discussion}

\noindent We first note that ForMIC performs exceptionally well on scenarios with infinite capacity resources and no obstacles, nearly equivalent to the performance of the centralized planner for all team sizes.
The introduction of depleting resources or obstacles gives either an agent allocation or path planning advantage to a centralized approach with full observability (but not real-time for larger teams).
With a limited FOV, ForMIC agents must continually search for new resources to exploit when foraging in an environment with depleting resources.
Nevertheless, ForMIC matches or exceeds the performance of all other decentralized algorithms across all team sizes.

We believe that high performance in these scenarios is the result of several key advantages of ForMIC.
First, handcrafting foraging strategies that rely on pheromones is difficult.
When pheromone trails from different agents overlap, many local minima and maxima form, making it hard to devise rules to follow specific trails or their gradients.
Unfortunately, this is especially true and problematic around the nest, since it is where agents must decide where to go next.
Our results suggest that ForMIC not only learns these complex pheromone patterns, but also learns to trust pheromone concentration data just the right amount.
The LSTM cell provides agents with implicit memory of their previous observations: encountered pheromone trails, resources, and more. 
In the case of obstacle-free environments with infinite resources, agents appear to rely largely on memory to navigate between resources and the nest; removing pheromones in these cases does not significantly decrease performance.
Differently, in dense environments with depleting resources, pheromones seem to be much more essential, since we observe a more significant drop in performance during and after wipeouts.
Nevertheless, across all of our tests, the removal of pheromones never seemed to stall ForMIC's foraging entirely, showing ForMIC's natural ability to handle a form of ``communication blackouts''.
Finally, the limited FOV enables agents to resolve local path planning conflicts: avoiding simple obstacles and other~agents.

Although C-SAF-11's agent movement rules cleverly construct a pheromone gradient, its reliance on predictable pheromone dynamics leads to the poor performance in the presence of pheromone disturbances/wipeouts.
Additionally, C-SAF-11 makes an unrealistic assumption for agent allocation at the nest: agents can see from the nest how much food is left at a resource by looking at the corresponding trail.
Despite not having this advantage, our approach outperformed C-SAF-11 (often very significantly).
Finally, C-SAF-11 does not have an approach for reallocating agents that reach the edge of the world before encountering a resource.

Cardinality-MR allows agents to exploit the closest resources to the nest and is also immune to wipeouts (since it does not rely on a pheromone model).
However, our results show that the efficiency of the food-gathering agents is not able to make up for the vast number of agents which are converted to beacons.
This leads to the algorithm often having the worst performance of those tested, especially for smaller teams.
This downside is mitigated in the largest teams (128 agents), where the small proportion of gathering agents still translates into a large amount of food deposited, while avoiding congestion around the nest.
In particular, and differently from C-SAF-11, we believe that Cardinality-MR would significantly benefit from a larger FOV for each agent, since fewer agents would be required to become beacons.


\section{Conclusion}
\label{RAL2022-conclusion}

\noindent In this paper, we leverage recent advances in distributed learning to propose a new avenue for pheromone-based MAF.
Our learned approach, ForMIC, endows agents with the ability to communicate implicitly by sensing local levels of pheromones dropped by other agents.
In a series of tests, we show that our approach outperforms state-of-the-art MAF algorithms, even in the presence of environmental dynamics that were not seen at training time. 
Additionally, the approach is shown to be scalable and performs well even in dense environments. 

Future work will extend ForMIC to more complex environmental dynamics.
One avenue will involve exploring environments with several classes or sizes of resources, where certain resources need to be manipulated and transported collaboratively by the agents.
Another research direction will aim at optimizing sustainability during foraging, by considering resources that regenerate with time unless totally exhausted (e.g., sustainable fishing). 
Finally, we will also investigate environments with more complex (e.g., concave) or dynamic obstacles, requiring advanced path planning.


\bibliographystyle{IEEEtran}
\bibliography{MAforaging}   

\end{document}